\pdfoutput=1

\documentclass[11pt]{article}

\usepackage[]{ACL2023}

\usepackage{times}
\usepackage{latexsym}

\usepackage[T1]{fontenc}

\usepackage[utf8]{inputenc}

\usepackage{microtype}

\usepackage{inconsolata}

\usepackage{amssymb}
\usepackage{bbm}
\usepackage{dsfont}
\usepackage{bbold}
\usepackage{mathbbol}
\usepackage{newtxmath}
\usepackage{booktabs}
\usepackage{multirow}
\usepackage{makecell}
\usepackage{graphicx}
\usepackage{subfigure}
\usepackage{listings}

\title{MSWA: Refining Local Attention with Multi-Scale Window Attention}

\author{Yixing Xu, Shivank Nag, Dong Li, Lu Tian \and Emad Barsoum \\
  Advanced Micro Devices, Inc., Beijing, China\\
  \texttt{\{yixing.xu, shivank.nag, d.li, lu.tian, emad.barsoum\}@amd.com} \\
}
\begin{document}
\maketitle

\begin{abstract}

Transformer-based LLMs have achieved exceptional performance across a wide range of NLP tasks. However, the standard self-attention mechanism suffers from quadratic time complexity and linearly increased cache size. Sliding window attention (SWA) solves this problem by restricting the attention range to a fixed-size local context window. Nevertheless, SWA employs a uniform window size for each head in each layer, making it inefficient in capturing context of varying scales.
To mitigate this limitation, we propose Multi-Scale Window Attention (MSWA) which applies diverse window sizes across heads and layers in the Transformer. It not only allows for different window sizes among heads within the same layer but also progressively increases window size allocation from shallow to deep layers, thus enabling the model to capture contextual information with different lengths and distances.
Experimental results on language modeling and common-sense reasoning tasks substantiate that MSWA outperforms traditional local attention in both effectiveness and efficiency.
\end{abstract}

\section{Introduction}

\begin{figure*}[ht] 

\centering 

\includegraphics[width=4.9in]{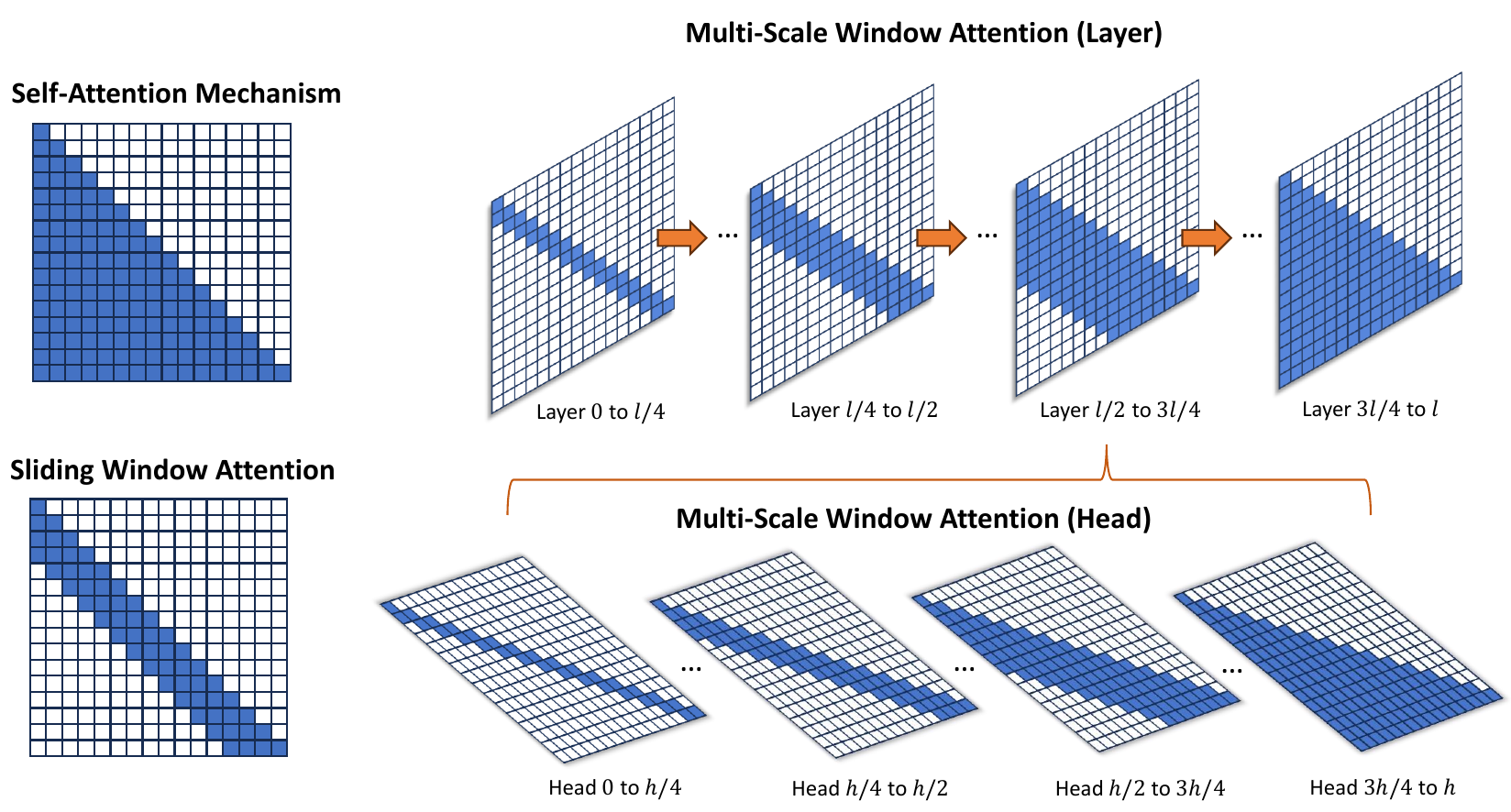} 

\caption{Illustration of Multi-Scale Window Attention mechanism.} \label{fig.main} 

\end{figure*}

The popularity of Transformer-based \cite{vaswani2017attention} large language models (LLMs) \cite{touvron2023llama,achiam2023gpt} has surged due to their remarkable performance on a wide range of applications, including NLP tasks like machine translation \cite{zhang2023prompting}, text summarization \cite{zhang2024benchmarking}, as well as more complex uses such as coding assistance \cite{ross2023programmer} and communicative agents \cite{li2023camel}. However, the standard Transformer employs the self-attention mechanism, whose quadratic time complexity becomes a bottleneck for the model's computational efficiency. Moreover, the KV cache required by the self-attention mechanism during inference autoregressively increases GPU memory consumption, making the deployment of LLMs unfriendly.

Recently, a lot of architectures have been proposed with the aim of being efficient foundations for LLMs \cite{gu2023mamba, peng2023rwkv, poli2023hyena}.  
One strand of research focuses on improving the efficiency of attention mechanism, such as sparse attention \cite{child2019generating}, sliding window attention \cite{beltagy2020longformer,zaheer2020big}, and linear attention \cite{choromanski2020rethinking,yang2023gated}.
While most studies focus on capturing global information of text sequence with linear computational time and limited memory,
sliding window attention (SWA) offers a more intuitive approach. By focusing on local information, it serves as a valuable mechanism for building LLMs~\cite{jiang2023mistral} or creating novel architectures~\cite{de2024griffin,arora2024simple}.



The key idea of SWA is to utilize the \textit{locality of reference} \cite{zaheer2020big} in NLP data, where most information about a token can be derived from its neighboring tokens. By allowing each token to attend to its neighbors within a fixed-size local window, SWA ensures linear computational complexity and constant KV cache consumption.
However, each head in every layer of the original SWA shares the same window size, ignoring the fact that the scale of contextual information can vary significantly.
For instance, a news report can span up to 2000 tokens, while a keyword might consist of only 4 tokens. Setting the attention window to the same size might lead to sub-optimal adaptation to contexts of different scales.
Additionally, different components of a Transformer model serve different roles. For example, shallower layers may exhibit more locality \cite{child2019generating}. Restricting all components to the same receptive field can severely impair the model's representation capacity.



To address the aforementioned issues, we propose a novel window attention variant called \textbf{M}ulti-\textbf{S}cale \textbf{W}indow \textbf{A}ttention (\textbf{MSWA}), which introduces diverse window sizes across heads and layers and improves the performance of SWA while reducing computation and memory cost.  
Specifically, we assign diverse window sizes to different heads within a layer to model contextual information at various lengths simultaneously. Moreover, we reduce the window size allocation for shallower layers and redistribute the resources to deeper layers, creating a pattern where shallow layers model local information and deep layers capture long-range dependencies.
We further propose an optional integration of MSWA with other efficient methods like linear attention, creating a model that is both local-sensitive and global-aware. Implementing MSWA on standard attention acceleration libraries \cite{dao2023flashattention} achieves efficiency beyond SWA without extensive additional development.

To validate the effectiveness of MSWA, we conduct extensive experiments. 
We train models from scratch for language modeling in various scenarios, including directly applying MSWA to the Transformer and combining MSWA with linear attention. Experimental results on word-level and character-level datasets demonstrate the superior language modeling ability of MSWA. Moreover, we verify the compatibility of MSWA to LLM by fine-tuning pre-trained LLM to adapt to the MSWA pattern. Performance on downstream common-sense reasoning tasks confirms the practical value of MSWA. We also conduct computational efficiency evaluation, where MSWA consistently achieves better efficiency compared to standard attention and SWA.

\section{Related Works} 

In this section, we briefly introduce the studies of large language models and attention mechanisms. 

\subsection{Large Language Models (LLMs)}

Language models \cite{bengio2000neural} have become a cornerstone of modern Natural Language Processing (NLP). Their primary purpose is to understand and generate human language, making them crucial for applications ranging from machine translation \cite{zhang2023prompting} to communicative agents \cite{li2023camel}. The advent of large-scale pre-trained models has significantly enhanced the performance of these applications.

Among them, Transformer-based models have revolutionized the field. Introduced by \citet{vaswani2017attention}, the Transformer architecture uses self-attention mechanism to process input sequences in a more parallelizable way. This innovation has led to the development of increasingly large and powerful models, such as GPT-4 \cite{achiam2023gpt}, Llama-3 \cite{llama3modelcard}, and Claude-3 \cite{anthropic2024claude}. These models, often termed "large language models", leverage vast amounts of data and computational resources to achieve state-of-the-art results on a wide array of NLP tasks.

\subsection{Attention Mechanisms}

The attention mechanism which enables the model to capture intricate dependencies across the entire sequence is at the heart of the Transformer's success. However, the standard self-attention mechanism has quadratic complexity with respect to the sequence length, which poses scalability challenges for longer sequences. To address this issue, many efficient attention variants have been proposed \cite{qiu2019blockwise,wang2020linformer,peng2021random,hua2022transformer}. For example, sliding window attention \cite{beltagy2020longformer,zaheer2020big,jiang2023mistral} limit attention to a fixed-size window around each token, making the computation more manageable for long texts. Linear attention methods \cite{choromanski2020rethinking,katharopoulos2020transformers,yang2023gated} approximate the attention calculation to reduce complexity from quadratic to linear.  These innovations have made it possible to apply Transformer to lengthy documents without prohibitive computational costs.




\section{Preliminaries}

In this section, we briefly introduce the preliminaries about self-attention, sliding window attention and linear attention operations.

\subsection{Self-Attention Mechanism} \label{attention}

For the Transformer models, its main computational and memory costs arise from the multi-head self-attention mechanism, focusing on two points: (1) quadratic time complexity over the input length, and (2) linearly increased size of the KV cache during inference. In the following, we analyze the attention mechanism based on the decoder form, as it is widely used in language models. 

Given input vectors $\{{\bf x}_i\}_{i=1}^{n} \in \mathbb R^D$, where each $x_i$ represents a single input token, $n$ is the sequence length, and $D$ is the dimension of the input vectors, a head in the self-attention layer first maps the input tokens into query vectors $\{{\bf q}_i\}_{i=1}^n$, key vectors $\{{\bf k}_i\}_{i=1}^n$, and value vectors $\{{\bf v}_i\}_{i=1}^n$:  
\begin{equation}  
{\bf q}_i = {\bf x}_i{\bf W}_q, ~~~ {\bf k}_i = {\bf x}_i{\bf W}_k,~~~ {\bf v}_i = {\bf x}_i{\bf W}_v.
\end{equation}  

where ${\bf W}_q$,  ${\bf W}_k$, ${\bf W}_v \in \mathbb R^{D\times d}$ are the mapping matrices, and $d$ is the dimension of each head. The output of this attention head is calculated by:


\begin{equation} \label{equa.1} 
    \alpha_{ij} = \frac{{\rm exp}({\bf q}_i {\bf k}_j^T / \sqrt{d})}{\sum_{t=0}^{i} {\rm{exp}}({\bf q}_i {\bf k}_t^T / \sqrt{d})}, 
\end{equation}

\begin{equation} 
    {\bf o}_i = \sum_{j=0}^{i} \alpha_{ij} {{\bf v}_j}. 
\end{equation} 

Performing the above computation for the entire sequence of length $n$ requires a time complexity of $O(dn^2)$ and a space complexity of $O(dn)$. 


\subsection{Sliding Window Attention} \label{swa}

Sliding Window Attention (SWA) is an efficient variant that restricts each token to attend to tokens within a local window of size $w$, as shown in Fig.~\ref{fig.main}. Given the query, key and value vectors ${\bf q}_i$, ${\bf k}_i$, ${\bf v}_i$, the output of the SWA head is defined as: 
\begin{equation} 
    \alpha_{ij} = \frac{{\rm exp}({\bf q}_i {\bf k}_j^T / \sqrt{d})}{\sum_{t=max(0,i-w)}^{i} {\rm{exp}}({\bf q}_i {\bf k}_t^T / \sqrt{d})}, 
\end{equation}

\begin{equation} 
    {\bf o}_i =  \sum_{j=max(0,i-w)}^{i} \alpha_{ij} {\bf v}_j. 
\end{equation} 

By using this method, the time and space complexity required for each head is reduced to $O(dnw)$ and $O(dw)$, respectively. 
Considering a Transformer with $l$ layers, each equipped with $h$ attention heads, the time and space complexity will become $O(dn(whl))$ and $O(d(whl))$ respectively.  
We can see that the computational and memory cost is proportional to $whl$, which is the summation of the window sizes of all the sliding window attention operations from all heads in all layers. 


\subsection{Linear Attention Mechanism} \label{linear}

Linear attention replaces the softmax operation in standard attention with a feature map-based dot product, eliminating the computation of $\rm{exp}(\cdot)$ and exchanging the matrix multiplication order, thereby achieving the goal of acceleration and constant memory cost. 
Specifically, given a kernel function $\phi(\cdot)$ that maps the $q_i$ and $k_j$ vectors into features, linear attention approximates ${\rm exp}({{\bf q}_i{\bf k}_j^T} / {\sqrt{d}})$ with the dot-product $\phi({\bf q}_i)\phi({\bf k}_j)^T$. Therefore, the output of an attention head is calculated by:

\begin{equation} 
    \alpha_{ij} = \frac{\phi({\bf q}_i)\phi({\bf k}_j)^T}{\sum_{t=0}^{i}\phi({\bf q}_i)\phi({\bf k}_t)^T}, 
\end{equation}

\begin{equation} 
    {\bf o}_i = \sum_{j=0}^{i} \alpha_{ij} {\bf v}_j = \frac{\phi({\bf q}_i)\sum_{j=0}^{i}\phi({\bf k}_j)^T{\bf v}_j}{\phi({\bf q}_i)\sum_{j=0}^{i}\phi({\bf k}_j)^T}. 
\end{equation} 

Based on the exchange of multiplication order, the computational time complexity of linear attention is reduced to $O(d^2n)$. In addition, during the auto-regressive inference process, both $\sum_{j=0}^{i}\phi({\bf k}_j)^T{\bf v}_j$ and $\sum_{j=0}^{i}\phi({\bf k}_j)^T$ can be written as a variable that is continuously accumulated, requiring only $O(d^2)$ space complexity.

\section{Multi-Scale Window Attention}

In this section, we present our proposed \textbf{M}ulti-\textbf{S}cale \textbf{W}indow \textbf{A}ttention (\textbf{MSWA}) mechanism, which leverages diverse window sizes across different heads and layers in Transformer architecture, as illustrated in Fig.~\ref{fig.main}.  
Our objective in designing this mechanism is to enhance the performance of SWA attention while maintaining computational and memory resources. Recall that the computational complexity and memory consumption required by SWA depend on the sum of the window sizes of all heads in all layers. 
On this basis, we not only change the distribution of window size allocation among different heads within the same layer, as introduced in Sec.~\ref{mswa-h}, but also adjust the distribution of window size allocation between layers, as detailed in Sec.~\ref{mswa-l}. The integration of these changes between heads and layers constitutes our MSWA mechanism, which will be introduced in Sec.~\ref{mswa}. Additionally, in Sec.~\ref{mswa-linear}, we provide an optional combination of the MSWA mechanism with the linear attention mechanism.
Implementation of MSWA can be found in Appendix \ref{app.impl}.


\subsection{Diverse Window Across Heads} \label{mswa-h}


This section focuses on dynamically changing the window size for each attention head within a layer. We refer to this mechanism as MSWA-h, as shown in the bottom right part of Fig.~\ref{fig.main}.

Different from SWA where all heads use the same window size $w_i$ in the $i$-th layer, in MSWA-h different heads have different scales of window sizes, and the summation of the total window sizes within a layer is less than that in the SWA, which is $w_ih$.
Specifically, inspired by many hierarchical architecture designs in the CV field \cite{liu2021swin}, we divide the attention heads into four groups and adjust the receptive field range with a $2\times$ change between each group, resulting in window sizes of $\frac{w_i}{4}$, $\frac{w_i}{2}$, $w_i$, $2w_i$, respectively.
Therefore, the summation of the total window sizes is: 
\begin{equation} 
    (\frac{w_i}{4} + \frac{w_i}{2} + w_i + 2w_i) \times \frac{h}{4} = \frac{15}{16}w_i h. 
\end{equation} 

Leveraging diverse window sizes among the heads within a layer allows the Transformer model to capture the relevant context of different scales simultaneously. 
This is because the outputs of different heads within an attention layer are concatenated together and then mapped through a matrix to form the final output of the layer, which allows contextual information at different distances to be integrated together. 
Additionally, considering the allocation of attention resources: all heads will attend to tokens within a distance of $\frac{w_i}{4}$ from the current token, while $\frac{3}{4}$ of the heads will attend to tokens within the $\frac{w_i}{4}$ to $\frac{w_i}{2}$ range, and so on. This implicitly models a long window with weighted emphasis, where the distribution of attention resources gradually decreases from near to far, aligning with the \textit{locality of reference} characteristic of text. 



\subsection{Diverse Window Across Layers} \label{mswa-l}

This section further introduces changing the allocation ratio of the attention window sizes between layers. We refer to this mechanism as MSWA-l, as illustrated in the upper right part of Fig. \ref{fig.main}.

To explain more clearly, we still use SWA as a comparison. In SWA, each attention layer has a total window size allocation of $hw$, where $h$ is the number of heads per layer, and $w$ is the base window size, which means that for any layer index $i$, $w_i = w$.  
In MSWA-l, the window size allocation varies across layers. More specifically, we divide all attention layers into several groups, and from shallow to deep, we continuously increase the total window size allocated to the attention layers in each group.
We adopt a similar setup to MSWA-h, with four groups having a $2\times$ change between each group, resulting in window size allocations of $\frac{hw}{4}$, $\frac{hw}{2}$, $hw$, and $2hw$, respectively.
The total window size resource allocated to all layers in MSWA-l is: 
\begin{equation} 
    (\frac{hw}{4} + \frac{hw}{2} + hw + 2hw) \times \frac{l}{4} = \frac{15}{16}whl. 
\end{equation}

For an attention layer, a larger window size allocation means that the window sizes of the heads in that layer are generally larger, allowing for the perception of a broader range of context. Therefore, gradually increasing the window size allocation from shallow to deep layers enables the model to focus on building local fine-grained information in the initial stages and progressively enhance the capture of long-distance relationships in the later stages. 
Additionally, the gradually expanding attention window allocation enables the model to continuously integrate local information from previous stages based on a larger receptive field.


\subsection{Integrate Diversity of Heads and Layers} \label{mswa}

\begin{table}[h!]
    \centering
    \resizebox{0.48\textwidth}{!}{
    \begin{tabular}{c|cccc}
        \toprule
       $i$ / $j$ & \textbf{$(0,\frac{h}{4}]$} & \textbf{$(\frac{h}{4},\frac{h}{2}]$} & \textbf{$(\frac{h}{2},\frac{3h}{4}]$} & \textbf{$(\frac{3h}{4},h]$} \\
        \midrule
        \textbf{$(0,\frac{l}{4}]$} & \( \frac{w}{16} \) & \( \frac{w}{8} \) & \( \frac{w}{4} \) & \( \frac{w}{2} \) \\
        \midrule
        \textbf{$(\frac{l}{4},\frac{l}{2}]$} & \( \frac{w}{8} \) & \( \frac{w}{4} \) & \( \frac{w}{2} \) & \( w \) \\
        \midrule
        \textbf{$(\frac{l}{2},\frac{3l}{4}]$} & \( \frac{w}{4} \) & \( \frac{w}{2} \) & \( w \) & \( 2w \) \\
        \midrule
        \textbf{$(\frac{3l}{4},l]$} & \( \frac{w}{2} \) & \( w \) & \( 2w \) & \( 4w \) \\
        \bottomrule
    \end{tabular} }
    \caption{Window size variation of MSWA with a base window size of \( w \). Here, \( i \) and \( j \) represent the indices of layer and head, respectively. Each element represents the \( w_{i,j} \) value when \( i \) and \( j \) are within a certain range.}
    \label{tab:window_sizes}
\end{table}

In this section, we aim to integrate the two strategies MSWA-h and MSWA-l introduced earlier to construct the final MSWA mechanism.

The description of MSWA starts with a base window size \( w \). In SWA, \( w \) is the window size used for all heads across all layers. In contrast, in MSWA, $w$ serves as the basis for window size variation.
We denote the base size value of the \( i \)-th layer as \( w_i \) and the actual window size of the \( j \)-th head in the \( i \)-th layer as \( w_{i,j} \). 
First, we evenly divide all attention layers into four groups. Depending on the group the layer belongs to, the values of $w_i$ from shallow to deep are $\frac{w}{4}$, $\frac{w}{2}$, $w$, $2w$, respectively. Further, within each layer \( i \), we divide all heads into four groups, each with different $w_{i,j}$ values, denoted as $\frac{w_i}{4}$, $\frac{w_i}{2}$, $w_i$ and $2w_i$.
Thus, the window size allocation for the entire Transformer model is: 
\begin{equation}
    \sum_{i=1}^l (\frac{w_i}{4} + \frac{w_i}{2} + w_i + 2w_i) \times \frac{h}{4} =  \frac{15}{16}\sum_{i=1}^l hw_i,
\end{equation}
which can be further derived as:
\begin{equation} 
    \frac{15}{16}(\frac{hw}{4} + \frac{hw}{2} + hw + 2hw) \times \frac{l}{4} \approx \frac{7}{8}whl. 
\end{equation}

The aforementioned window variation method is demonstrated in Tab. \ref{tab:window_sizes}.
Note that MSWA can benefit from the advantages of both MSWA-h and MSWA-l. When computing within a single layer, it can capture both long-range and short-range contextual information at the same time, and allocate different attention resources to information at various distances. When transitioning the information from one layer to another, MSWA continuously enhances the overall perception scope, integrating previous local information into a broader synthesis. 



\subsection{Combination with Linear Attention} 
\label{mswa-linear}

\begin{figure}[ht] 

\centering 
\includegraphics[width=3.0in]{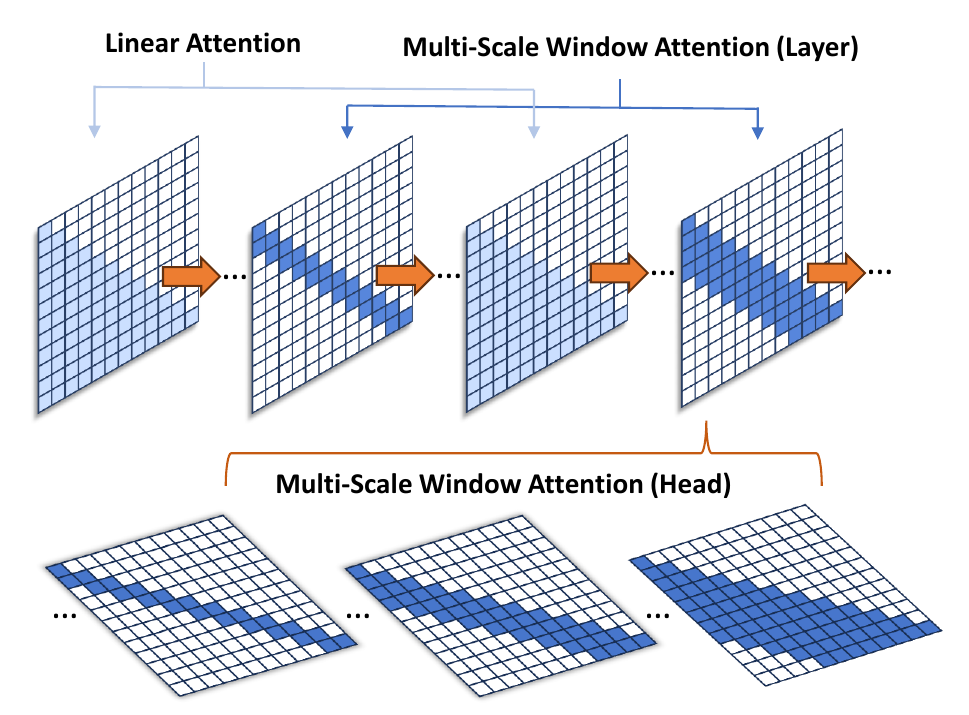} 

\caption{ Combination of MSWA and linear attention.} \label{fig.linear} 

\end{figure}

This section further proposes combining MSWA with an efficient global attention mechanism, \textit{i.e.,} linear attention, as illustrated in Fig. \ref{fig.linear}.

As we introduced earlier, many efficient mechanisms focus on capturing global information with limited resources. Therefore, they often fail to allocate high importance to relevant local information. Linear attention is a typical example of this issue. As introduced in Sec.~\ref{linear} it stores global sequence information in fixed-size variables $\sum_{j=0}^{i}\phi({\bf k}_j)^T{\bf v}_j$ and $\sum_{j=0}^{i}\phi({\bf k}_j)^T$, which can lead to a loss of attention focus \cite{qin2022devil}.


Therefore, we propose to combine MSWA with linear attention to compensate for its shortcoming and achieve a balance between efficiency and performance. Specifically, we alternately stack MSWA layers and linear attention layers. For example, the $l$ layers in a combined model  are evenly divided into four groups, each contains $\frac{l}{12}$ linear attention layers and $\frac{l}{6}$ MSWA layers stacked together. For all MSWA layers in the entire model, we consider them as a whole and utilize the same window size variation method introduced in Sec.~\ref{mswa} and adjust the window sizes across layers and heads.

\section{Experiments}

\begin{table*}[ht]
  \centering
    \resizebox{0.8\textwidth}{!}{
    \begin{tabular}{c|c|c|c|cc}
    \toprule
    \multicolumn{2}{c|}{\textbf{Attention Mechanism}} & \textbf{Length Setting} & \textbf{Relative Cost $\downarrow$} & \makecell{\textbf{Wikitext-103} \\ PPL $\downarrow$} & \makecell{\textbf{enwik8} \\ bpc $\downarrow$} \\
    \midrule
    \multicolumn{2}{c|}{\multirow{2}[1]{*}{Standard Self-Attention}} & $n=1024$ & \textbf{9.10} & 28.61 & 1.12 \\
    \multicolumn{2}{c|}{} & $n=2048$ & 18.20  & \textbf{28.33} & \textbf{1.10} \\
    \midrule
    \multirow{4}[1]{*}{\makecell{Local \\ Attention}} & SWA  & $w=128$ & 1.14  & 30.70  & 1.22 \\
          & MSWA-h & $w_{i,j}$ from 32 to 256 & 1.07  & 29.96 & 1.16 \\
          & MSWA-l & $w_{i,j}$ from 32 to 256 & 1.07  & 30.19 & 1.16 \\
          & MSWA  & $w_{i,j}$ from 8 to 512 & \textbf{1.00} & \textbf{29.56} & \textbf{1.11} \\
    \bottomrule
    \end{tabular} }
      \caption{Evaluation of directly applying each attention mechanism as Transformer backbone for language modeling. For length settings, $n$ represents the sequence length for standard self-attention, $w$ represents the window size for SWA, and $w_{i,j}$ represents the window size of the $j$-th head in the $i$-th layer in the MSWA series. Relative cost measures the computational and memory cost of each mechanism as a proportion of the cost of MSWA, with the ratio being consistent for both types of complexity ($\frac{dn^2}{dnw} = \frac{dn}{dw}$).}
  \label{tab:1.1}%
\end{table*}%

This section demonstrates the effectiveness of our MSWA mechanism. 
An overview of experiments, the main datasets and the baselines are described below. 
More experimental details, including the implementation dependencies and the detailed setup for each experiment, are shown in the Appendix \ref{app.exp}.

\paragraph{Overview.} 

Sec.~\ref{exp.1} presents language modeling evaluation on natural language datasets in two different scenarios. In Sec.~\ref{exp.1.1}, we directly applying MSWA and its sub-mechanisms to the Transformer model. In Sec. \ref{exp.1.2}, we combine the MSWA mechanism with the linear attention mechanism.
Sec.~\ref{exp.2} verifies the compatibility of MSWA with existing LLM in downstream tasks. We fine-tune the pre-trained Llama2-7B \cite{touvron2023llama} model to adapt to new attention patterns, followed by few-shot evaluations on a series of common-sense reasoning tasks. 
Additionally, in Sec.~\ref{exp.speed}, we compare the computational efficiency of MSWA with other attention mechanisms. Sec.~\ref{exp.ablation} provides a series of ablation experiments.

\paragraph{Datasets.}  

We use both word-level and character-level natural language datasets for language modeling evaluation, specifically Wikitext-103 \cite{merity2016pointer} and enwik8 \cite{mahoney2011large}. Wikitext-103 is a word-level language modeling benchmark containing over 100M tokens, while enwik8 is a character-level dataset consisting of 100M bytes, both originally sourced from Wikipedia text. 
For the evaluation on downstream tasks, we use the RedPajama \cite{together2023redpajama} dataset for fine-tuning and perform downstream few-shot evaluation on eight common-sense reasoning benchmarks: PIQA \cite{bisk2020piqa}, OpenBookQA \cite{mihaylov2018can}, WinoGrande \cite{sakaguchi2021winogrande}, HellaSwag \cite{zellers2019hellaswag}, BoolQ \cite{clark2019boolq}, COPA \cite{roemmele2011choice}, ARC easy and challenge \cite{clark2018think}.

\paragraph{Baselines.} We mainly compare MSWA with two baseline methods: 
1) {Standard Self-Attention}: We use the standard self-attention mechanism as a strong baseline. As introduced in Sec.~\ref{attention}, it can attend to all tokens in the whole sequence, achieving excellent performance at the cost of quadratic time and linear space complexity. 
2) {Sliding Window Attention}: SWA is the most widely used variant of local attention, with applications including direct construction of LLMs~\cite{jiang2023mistral}, integration with global architectures \cite{de2024griffin,arora2024simple}.  
As described in Sec.~\ref{swa}, it only attends to tokens within a fixed window size, thereby save time and space costs. 





\subsection{Language Modeling Evaluation} \label{exp.1}

In this section, we evaluate the language modeling capabilities of MSWA mechanism by training models from scratch on Wikitext-103 and enwik8. 




\begin{table}[tbp!]
  \resizebox{0.5\textwidth}{!}{
    \begin{tabular}{c|c|cc}
    \toprule
    {\textbf{Architecture}} & {{\makecell{\textbf{Relative} \\ \textbf{Cost} $\downarrow$}}} &
{{\makecell{\textbf{Wikitext-103} \\ PPL $\downarrow$}}} & {\makecell{\textbf{enwik8} \\ bpc $\downarrow$}} \\
    \midrule
    Transformer & 18.20 &  \textbf{29.43} & 1.15 \\
    \midrule
    Linear Attention & 0.69 & 40.57  &  1.29 \\
    Linear Attention + SWA  & 0.98 &31.85 & 1.16 \\
    Linear Attention + MSWA  & 0.89 & 30.83 & {1.13} \\
    MSWA  & 1.00  & 30.38 & \textbf{1.12} \\
    \bottomrule
    \end{tabular} }
      \centering
        \caption{Evaluation of combining MSWA with Linear Attention mechanism.
        Note that the setup for this experiment differs from that in Table \ref{tab:1.1}, resulting in different performance. For detailed experimental settings, please refer to Appendix \ref{setup.exp-1}.
        }
  \label{tab:1.2}%
\end{table}%

\begin{table*}[htbp!]
  \centering
  \resizebox{0.9\textwidth}{!}{
    \begin{tabular}{c|ccccccccc}
    \toprule
    \textbf{Attention} & \textbf{PIQA} & {\textbf{OBQA}} & \textbf{WinoGrande} & \textbf{HellaSwag} & {\textbf{BoolQ}} & \textbf{COPA} & \textbf{ARC-e} & {\textbf{ARC-c}} & \textbf{Average} \\
    \midrule
    \multicolumn{10}{l}{\textit{Performance under 3-Shot setting}} \\
    SWA   & 57.56  & \textbf{28.00}  & \textbf{54.93}  & 45.21  & \textbf{68.07}  & \textbf{61.00}  & 36.24  & 25.94 & 47.12 \\
    MSWA  & \textbf{66.10}   & 24.60 & 51.14  & \textbf{55.44}  & 67.52 & 57.00  & \textbf{42.38}   & \textbf{27.99} & \textbf{49.02} \\
    \midrule
    \multicolumn{10}{l}{\textit{Performance under 5-Shot setting}} \\
    SWA   & 56.64  & \textbf{29.40}  & 50.59  & 44.40  & 55.87  & 48.00  & 32.79  & 23.63  & 42.66 \\
    MSWA  & \textbf{67.03}  & 28.80 & 
    \textbf{51.85}  & \textbf{61.78}  & \textbf{56.27} & \textbf{56.00}  & \textbf{46.93}    & \textbf{30.46} & \textbf{49.89} \\
    \bottomrule
    \end{tabular}%
    }
      \caption{Few-shot accuracy results (\%) on common-sense reasoning tasks for Llama-7B, after fine-tuned in the RedPajama dataset with each attention mechanism.}
  \label{tab:shot}%
\end{table*}%

\subsubsection{Direct Construction of Language Model} \label{exp.1.1}

This section presents the results of directly using MSWA as the Transformer backbone, which is a straightforward way to validate its performance. 

As shown in Tab. \ref{tab:1.1}, we report perplexity (PPL) results on the Wikitext-103 test set and bits-per-character (bpc) results on the enwik8 test set. 
The experimental results demonstrate that:  
1) MSWA achieves better language modeling performance compared to SWA with smaller computational and memory cost, reducing PPL by 1.14 on Wikitext-103 and bpc by 0.11 on enwik8. 
2) Dynamically adjusting the window size from either the layer or the head perspective results in improved language modeling capability, and combining both approaches yields further enhancements. 
3) Although there is still a performance gap between local attention and the standard self-attention, MSWA can achieve closer or similar results to standard attention. For example, on enwik8, MSWA obtains a bpc that is 0.01 lower compared to standard attention with a sequence length of 1,024 and 0.01 higher compared to a sequence length of 2,048, while requiring significantly fewer resources than both.

\subsubsection{Combination with Linear Attention} \label{exp.1.2}

This section demonstrates the combination of MSWA and Linear Attention mechanism, achieving language modeling capabilities comparable to the standard Transformer in a more efficient way. We use the 2nd-order Taylor series feature map \cite{zhang2023hedgehog,arora2024simple} as the kernel function for linear attention.

The experimental results are shown in Tab. \ref{tab:1.2}. We can conclude that: 
1) Combining MSWA with linear attention achieves comparable performance to the standard Transformer. Specifically, on Wikitext-103 the combined model achieves a PPL that is only 1.4 higher than the Transformer, while on enwik-8, it achieves a 0.2 lower bpc compared to the Transformer. 
2) The performance of linear attention is greatly improved when combined with either SWA or MSWA. Among them, combining with MSWA yields better language modeling performance, providing direction for future researches. 
3) Compared to directly using MSWA, combining MSWA with linear attention achieves a balance between performance and efficiency, enhancing efficiency with minimal loss in performance.

\subsection{Evaluation on Downstream Tasks} \label{exp.2}

\begin{figure*}[ht] 

\centering 

\subfigure[Base window size $w=128$.]{ 

\includegraphics[width=2.0in]{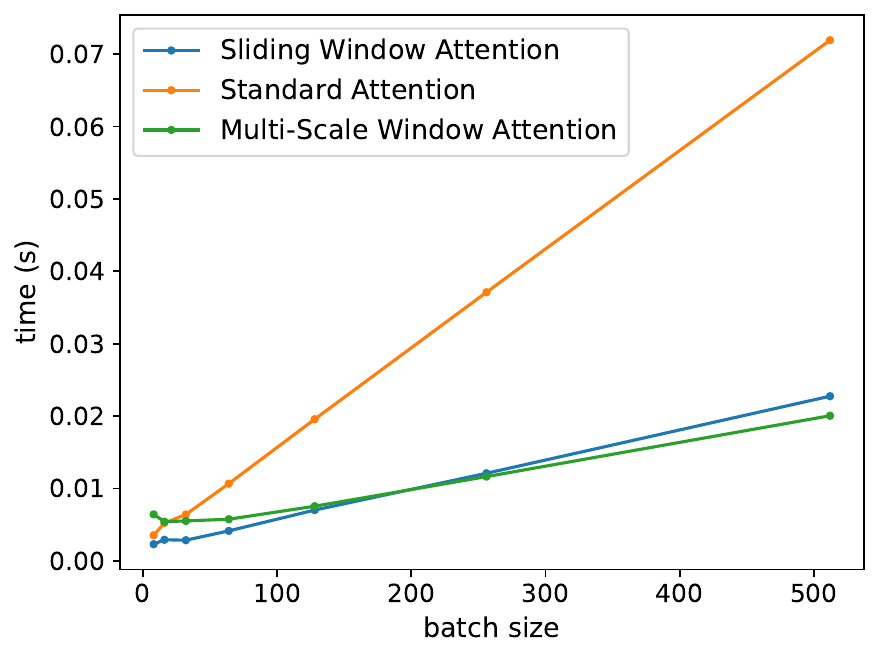} 


}%
\subfigure[Base window size $w=256$.]{ 
\includegraphics[width=2.in]{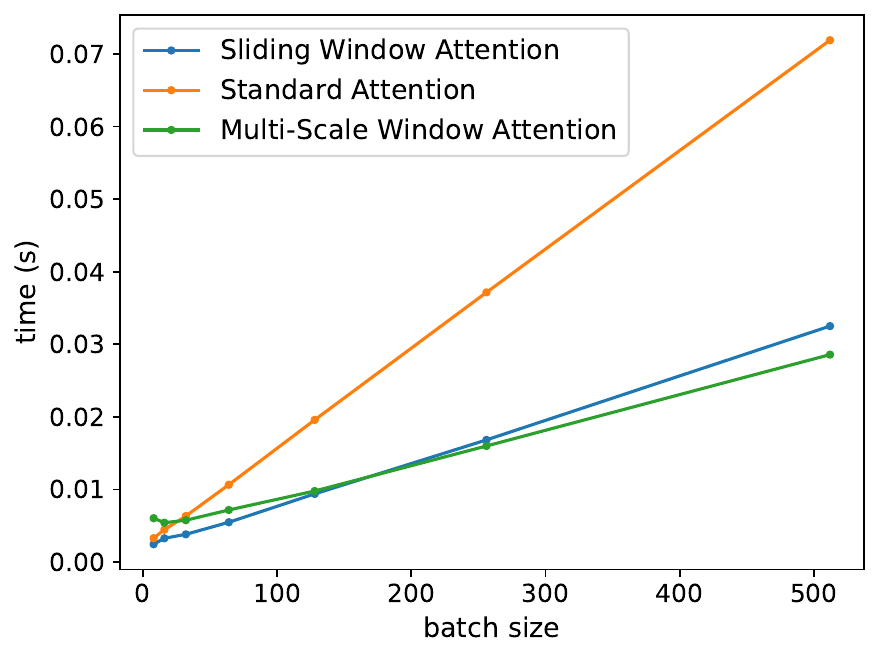} 
}%
\subfigure[Base window size $w=512$.]{ 
\includegraphics[width=2.0in]{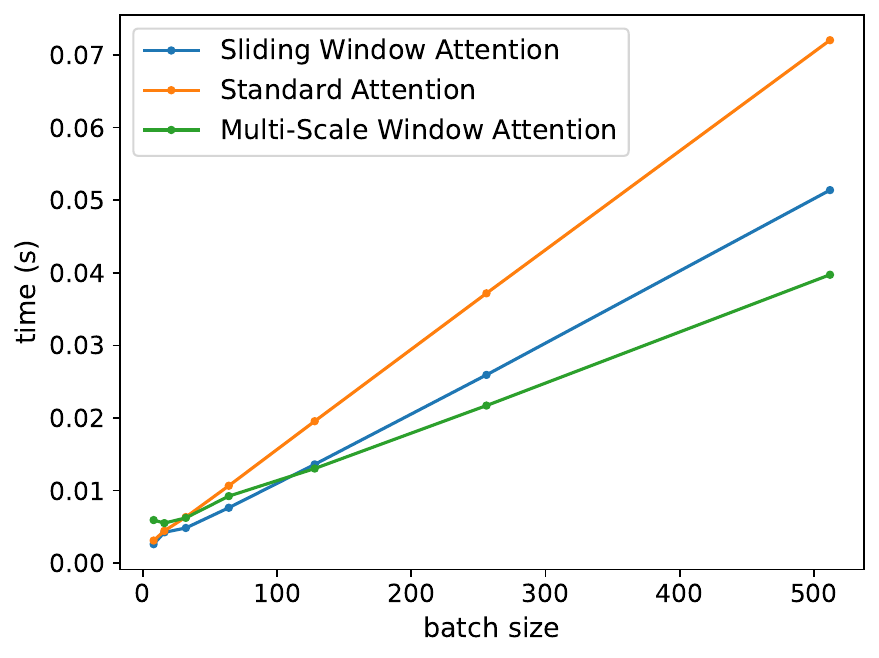} 
}%

\caption{ Computational time required by each attention mechanism to predict the next token.} \label{fig.time} 

\end{figure*}

In this section, we evaluate the performance of Llama-7B after fine-tuning with local attention patterns and testing on downstream common-sense reasoning tasks. The purpose of this evaluation is to verify the compatibility of MSWA with the current pre-trained LLM and its effectiveness when scaled to a large number of model parameters.

Tab. \ref{tab:shot} presents the accuracy results of the models on various downstream benchmarks using 3-shot and 5-shot settings. The experimental results indicate that: 
1) The MSWA mechanism demonstrates better common-sense reasoning ability compared to traditional sliding window attention, with average accuracy differences of $+1.9$ and $+7.23$ in 3-shot and 5-shot scenarios, respectively. 2) The MSWA mechanism shows a stronger ability to adapt to different context lengths, with its performance remaining stable across varying shot numbers, whereas SWA's average accuracy decreases by 4.46 as the number of shots increases.

\subsection{Computational Efficiency Evaluation} \label{exp.speed}

This section further evaluates the actual computational efficiency of our MSWA mechanism. Unlike the inferable size of the KV cache, the computational speed of Transformer models needs to be measured in practice to obtain realistic results. Therefore, we measure the time required to predict the next token during forward propagation in inference process for various attention mechanisms. 
We utilize FlashAttention \cite{dao2023flashattention} for the computation of various attention mechanisms.


The computational efficiency is shown in Fig. \ref{fig.time}. From the experimental results, we can observe the following: 
1) Both SWA and MSWA has a significant efficiency advantage compared to the standard self-attention mechanism. This advantage becomes more pronounced as the batch size increases. 
2) Compared to the traditional SWA, MSWA adapts better to larger batch sizes, often achieving better performance as the batch size increases. 
3) For larger base window sizes, the efficiency advantage of MSWA becomes even more apparent, making it well-suited for scaling up context lengths.

\subsection{Ablation Study} \label{exp.ablation}

We conduct a series of ablation studies in this section, mainly focusing on the impact of the base window size on the MSWA mechanism, as well as the effects of other window variation strategies on the MSWA mechanism. 
In these experiments, we use the same setup from Sec.~\ref{exp.1.1}, only alter the sizes of windows and the variation method.

\subsubsection{Effect of Base Window Size}

The impact of the base window size on the MSWA mechanism is shown in Tab. \ref{tab:ablation.1}, where for each $w$, the window size variation method for MSWA is introduced as in Sec.~\ref{mswa}. It can be observed that: 1) In each case from $w = 64$ to $w = 512$, MSWA achieves better results compared to traditional local attention SWA. 2) MSWA can achieve better performance than traditional local attention with less than half the resource consumption. For example, SWA with a window size of 512 achieves a PPL of 29.20 on Wikitext-103, while MSWA evolved from a base window size 256 achieves a PPL of 28.92.

\begin{table}[htbp]
  \centering
    \resizebox{0.48\textwidth}{!}{
    \begin{tabular}{c|c|c|c}
    \toprule
    \textbf{Attention} & \textbf{Length Setting} & \makecell{\textbf{Relative} \\ \textbf{Cost $\downarrow$}} & \makecell{\textbf{Wikitext-103} \\ PPL $\downarrow$} \\
    \midrule
    SWA   & $w=512$ & 4.55 & 29.20 \\
    MSWA  & $w_{i,j}$ from 32 to 2048 & 4.00 & \textbf{28.67} \\
    \midrule
    SWA   & $w=256$ & 2.28 & 29.93 \\
    MSWA  & $w_{i,j}$ from 16 to 1024 & 2.00 & \textbf{28.92} \\
    \midrule
    SWA   & $w=128$ & 1.14 & 30.70 \\
    MSWA  & $w_{i,j}$ from 8 to 512 & 1.00 & \textbf{29.56} \\
    \midrule
    SWA   & $w=64$ & 0.57 & 31.90 \\
    MSWA  & $w_{i,j}$ from 4 to 256 & 0.50 & \textbf{30.35} \\
    \bottomrule
    \end{tabular} }
      \caption{Ablation on scaling base window size.
      In every case where $w$ ranges from 64 to 512, MSWA with a smaller cost achieves better results compared to SWA.}
  \label{tab:ablation.1}%
\end{table}%

\subsubsection{Effect of Window Variation Strategy}

The comparative results with other window variation strategies are shown in Tab. \ref{tab:ablation.2}. We consider two approaches. 
In the first approach, to demonstrate the effectiveness of our layer-wise allocation, where lower layers model local information and higher layers capture long-range information, we reverse the original window size allocation between layers, which means reducing the window size from shallow to deep layers.
In the second approach we change the window size variation between each group from multiplying by 2 each time to an arithmetic progression. For example, for the base window size of 128, we change the evolution of each group to \{64, 96, 128, 160\}. The experimental results demenstrate that the performance achieved by both variation strategies is slightly weaker than the method we introduced previously. 


\begin{table}[htbp]
  \centering
    \resizebox{0.42\textwidth}{!}{
    \begin{tabular}{c|c}
    \toprule
    \textbf{Variation Strategy} & \makecell{\textbf{Wikitext-103} PPL $\downarrow$} \\
    \midrule
    Ours  & \textbf{29.56} \\
    Decreasing for Deeper Layer & 30.46 \\
    Arithmetic Progression & 29.90 \\
    \bottomrule
    \end{tabular} }
      \caption{Ablation on window variation strategies.
      "Decreasing for Deeper Layer" refers to reversing MSWA layer-wise window allocation by decresing the window size from shallow to deep layers. "Arithmetic Progression" means changing the window size variation among different groups to an arithmetic increase.
      }
  \label{tab:ablation.2}%
\end{table}%

\section{Conclusion} 

We propose a novel window attention variant called \textbf{M}ulti-\textbf{S}cale \textbf{W}indow \textbf{A}ttention (\textbf{MSWA}), which leverages diverse window sizes for different heads in different layers. Compared to the traditional sliding window attention, which is inefficient in capturing context of varying scales, we enable the model to capture contextual information of varying lengths and distances with less computational resources and memory usage.
Experimental results on lanaguage modeling and common-sense reasoning tasks demonstrate that MSWA can outperform previous local attention mechanism, while obtaining better efficiency.

\clearpage


\bibliography{custom} 

\bibliographystyle{acl_natbib}

\appendix 

 \clearpage

\section{Implementation of MSWA} 
\label{app.impl}

Here, we provide the specific implementation of MSWA mechanism.
Essentially, MSWA is a combination of two sub-mechanisms: MSWA-h and MSWA-l. Their implementations in the program are independent of each other.

 \subsection{Implementation of MSWA-h}
 
As for the implementation of MSWA-h, the overall flow is consistent with the standard attention layer in Transformer, except for the grouped implementation of the multi-head attention. 
Since different groups use different window sizes, we first use the \textit{reshape} function from PyTorch \cite{paszke2019pytorch} to divide all head's $\bf q$, $\bf k$, and $\bf v$ vectors into different groups. For heads within the same group, we use the efficient methods of previous SWA implementations (\textit{e.g.} FlashAttention \cite{dao2023flashattention}, xFormers \cite{xFormers2022}) for parallel computation. Calculations between different groups are carried out separately. After completing the attention calculation of all groups, we use \textit{cat} function of PyTorch to concatenating the attention outputs of each group together in the group's dimension, and then project them onto the final output of this attention layer through a matrix. Therefore, we can implement the MSWA-h mechanism without much additional development.


\subsection{Implementation of MSWA-l}

Regarding the implementation of MSWA-l, it is simpler compared to MSWA-h because in the Transformer model, different layers are stacked, and the computations between different layers are sequential and completely independent. We only need to assign the window size allocation for each layer as a parameter and pass it to the initialization function of each layer object.

\section{Experimental Details} \label{app.exp}

\subsection{Dependencies}

For all the methods in our experiments, we implement them using the PyTorch \cite{paszke2019pytorch} and FlashAttention \cite{dao2023flashattention} library. Additionally, the training and test process for the experiments in Sec.~\ref{exp.1} is based on Fairseq \cite{ott2019fairseq}, while for the experiments in Sec.~\ref{exp.2}, the fine-tuning process is based on DeepSpeed \cite{rasley2020deepspeed} and the evaluation process is implemented using lm-evaluation-harness \cite{eval-harness}.  
The efficiency evaluation in Sec.~\ref{exp.speed} is performed on a NVIDIA A100 GPU. 

\subsection{Setups for Each Experiment}
In this section, we introduce the setups and training details for each experiment.
\subsubsection{Language Modeling Evaluation} \label{setup.exp-1}

\paragraph{Direct Construction of Language Model} 
For all attention mechanisms, we apply them to a 12-layer standard Transformer model, with each layer having 8 attention heads. The model dimension and head dimension are 512 and 64, respectively.  
To better simulate the model's operation on the long-range sequence and reflect the memory overhead of various mechanisms, we introduce the cache mechanism from Transformer-XL \cite{dai2019transformer} and simultaneously adopt its relative position embedding. 
Each model is trained from scratch on two datasets based on the casual language modeling objective for 150,000 update steps. 
The number of tokens trained per step, which is the product of batch size and sequence length, is kept consistent (16,384 for Wikitext-103, 49,152 for enwik8). 
We use the AdamW \cite{loshchilov2018decoupled} optimizer with beta values of (0.9, 0.98), set the learning rate to 2e-4 with 1,000 warm-up steps, and use a cosine learning rate scheduler.

\paragraph{Combination with Linear Attention} For the linear attention mechanism, we use the 2nd-order Taylor series feature map \cite{zhang2023hedgehog,arora2024simple} as the kernel function. Following the setup by \citet{arora2024simple}, we employ RoPE \cite{su2021roformer} encoding for both linear attention and local attention, and use RMSNorm and SwiGLU mechanism.  
Each model consists of 12 layers. For combination of linear attention and local attention, each consecutive stack of three layers containing one linear attention layer and two local attention layers.  
The model dimension, head dimension, and feature dimension are set to 512, 64, and 16, respectively.  
The base window size for local attention is 128. 
During training and evaluation, the data is segmented into sequences containing 2,048 tokens without using the Transformer-XL style caching mechanism.  
The batch size for both datasets is 8. Other training settings are the same as in Sec.~\ref{exp.1.1}.

\subsubsection{Evaluation on Downstream Tasks} \label{setup-exp.2}

For the fine-tuning process, the Llama-7B model is trained for 2,000 steps using each local attention pattern based on the casual language modeling objective. The global batch size for each step is 32, and each sample consists of a sequence of 4,096 tokens.  
We use the AdamW optimizer with beta values of (0.9, 0.95). After 20 warm-up steps, the learning rate is fixed at 2e-5. We apply the LoRA \cite{hu2021lora} technique with $r=8$ and $\alpha = 16$ to train the attention parameters. Inspired by \citet{chen2023longlora}, we also make the normalization and embedding layers trainable.  
During fine-tuning and downstream testing, for SWA we set the base window size $w$ to $1/4$ of the sequence length. For the MSWA series, the window size dynamically evolves based on $w$, ensuring consistent resource usage with SWA.

\subsubsection{Computational Efficiency Evaluation}
\label{setup-exp.3}

For all attention mechanisms, their computation is based on the FlashAttention library, which is the current standard method for efficiently implementing attention operation. 
We apply them in a 32-layer Transformer model, with each layer containing 16 attention heads. The model dimension and head dimension are 1,024 and 64, respectively. We use a sequence of 2,048 tokens for measurement and report the median value across the computation time at positions \{500, 1,000, 1,500, 2,000\} in the sequence. The batch size is set to \{8, 16, 64, 128, 256, 512\}, and we record the experimental results for each case.

\end{document}